\documentclass{article}

\PassOptionsToPackage{numbers, compress}{natbib}

\usepackage[preprint]{neurips_2026}

\usepackage[utf8]{inputenc}
\usepackage[T1]{fontenc}
\usepackage{hyperref}
\usepackage{url}
\usepackage{booktabs}
\usepackage{amsfonts}
\usepackage{nicefrac}
\usepackage{microtype}
\usepackage{graphicx}
\usepackage{natbib}
\usepackage{multirow}
\usepackage{adjustbox}
\usepackage{xcolor}
\usepackage{colortbl}
\usepackage{amsmath,amssymb}
\usepackage{subcaption}
\usepackage{wrapfig}
\usepackage{enumitem}
\graphicspath{{media/}}
\usepackage{comment}

\definecolor{posblue}{HTML}{1A6FB5}
\definecolor{negred}{HTML}{C0392B}
\definecolor{lightgray}{gray}{0.92}

\title{Dual-Pathway Circuits of Object Hallucination in Vision-Language Models}

\author{%
Jiaxin Liu\thanks{Corresponding authors: Jiaxin Liu (\texttt{jiaxin26@illinois.edu}) and Guangyuan Dong (\texttt{guangyuan@u.nus.edu}).} \\ UIUC \\
  \And Ding Zhong \\ UMich \\
  \And Yue Wang \\ Stanford \\
  \AND Zhidong Yang \\ HKUST \\
  \And Zhaolu Kang \\ PKU \\
  \And Guangyuan Dong\footnotemark[1] \\ NUS \\
  \AND Qishi Zhan \\ Marquette \\
  \And Pengcheng Fang \\ Southampton \\
  \And Aofan Liu \\ PKU \\
}

\begin{document}
\maketitle


\begin{abstract}
Vision-language models (VLMs) have demonstrated remarkable capabilities in bridging visual perception and natural language understanding, enabling a wide range of multimodal reasoning tasks. 
However, they often produce object hallucinations, describing content absent from the input image, which limits their reliability and interpretability.
To address this limitation, we propose Dual-Pathway Circuit Analysis, a framework that identifies and characterizes hallucination-related circuits in VLMs for mechanistic understanding and causal probing.
We first apply activation patching across five architecturally diverse VLMs to identify a visual grounding pathway that supports correct predictions and a hallucination pathway that drives erroneous outputs. 
We then introduce Conditional Pathway Analysis (CPA) to characterize pathway-level interactions, revealing that grounding components remain strongly redundant in both correct and hallucinating samples but undergo a consistent polarity flip, shifting from supporting the ground truth on correct samples to aligning with the hallucinated answer on erroneous ones. 
We further perform targeted suppression of hallucination-pathway components, showing that scaling these components reduces object hallucination by up to 76\% with minimal accuracy cost, and validate that the same circuit selectively transfers to relational but not attribute hallucination.
Evaluations on POPE-adversarial and AMBER show that the identified circuits are consistent across architectures, support causal intervention, and transfer selectively across hallucination types. Code is available at \url{https://github.com/jiaxin26/DualPath-VLM}.



\end{abstract}

\section{Introduction}
\label{sec:intro}
Vision-Language Models (VLMs) demonstrate strong visual reasoning~\citep{yuan2025video} and cross-modal understanding~\citep{feng2025videor1,shen2025vlmr1}. Representative systems such as Qwen-VL~\citep{bai2025qwen3,bai2025qwen2} and GPT-4~\citep{openai2024gpt4technicalreport} consequently deliver strong performance on image captioning~\citep{zhao2025imagecaption}, visual grounding~\citep{bai2025univgr1}, and video understanding~\citep{zhang2024mme}.
However, VLMs often produce object hallucinations, describing entities, attributes, or relations absent from the input image. Such outputs reflect over-reliance on text-centric priors rather than visual evidence, limiting model reliability and interpretability. Among hallucination types, object-existence questions provide a particularly clean setting: their binary ground truth directly tests whether the model grounds its answer in visual evidence, while avoiding many confounds introduced by attribute or relational reasoning. This makes them a natural entry point for causal probing of hallucination-related computations. Benchmarks including POPE~\citep{li2023pope} and CHAIR~\citep{rohrbach2018chair} systematically document this problem. Existing mitigation strategies span training-time alignment~\citep{sun2023aligning,yu2024rlhfv}, decoding-time constraints~\citep{huang2024opera,leng2024vcd}, and post-hoc corrections~\citep{yin2023woodpecker}. These approaches primarily treat VLMs as black boxes, modifying input signals or decoding procedures without investigating the internal computations behind hallucination. As a result, these methods are often model-specific and generalize poorly across architectures.
 
Mechanistic interpretability offers a complementary perspective, asking not only whether hallucinations occur but how they arise within the network. Prior work has used activation patching to identify circuits underlying behaviors in language models, including indirect object identification~\citep{wang2023interpretability}, factual recall~\citep{meng2022locating}, and truthfulness~\citep{li2024inference}. Studying hallucination in VLMs presents additional challenges, as the relevant computations may span both visual representations and language-model components, and modern VLMs differ substantially in how visual information is integrated. This raises a fundamental question: are hallucination-related circuits architecture-specific, or do they reflect a more general property of VLM training?

To address this question, we propose a cross-model analysis strategy to fully investigate the mechanism of object-existence hallucination on POPE-adversarial across five architecturally diverse VLMs. Specifically, our analysis proceeds in three stages. We first apply activation patching to localize hallucination-related components, revealing a shared dual-pathway organization across all five models. We then introduce Conditional Pathway Analysis (CPA) to characterize how components within each pathway interact, exposing a polarity-flip signature invisible to standard component-level analysis. Finally, we use targeted suppression as a causal test of the circuit hypothesis and evaluate whether the identified pathway transfers across hallucination types. Together, these results are consistent with object hallucination in VLMs having a shared mechanistic structure that develops during training rather than being inherited from a specific architecture.
In summary, our contributions are three-fold:
\begin{itemize}[leftmargin=*,nosep]
\item \textbf{Cross-model dual-pathway discovery:} We apply activation patching across five architecturally diverse VLMs and identify a consistent dual-pathway organization, consisting of a visual grounding pathway and a hallucination pathway which may lead to errors (Section~\ref{sec:circuits}).
\item \textbf{Pathway-level interaction analysis:} We propose Conditional Pathway Analysis (CPA) to quantify within-pathway interactions, discovering a consistent grounding-pathway IE polarity flip between correct and hallucinating samples that is invisible to single-component patching (Section~\ref{sec:cpa_results}).
\item \textbf{Causal validation and benchmark transfer:} Component-wise suppression of the hallucination pathway reduces object hallucination by up to 76\% with $\leq$2\,pp accuracy cost. Matched static-direction baselines are consistently less effective than per-component scaling, and the identified circuit transfers to relational but not attribute hallucination on AMBER (Sections~\ref{sec:intervention_results}--\ref{sec:cross_benchmark}).
\end{itemize}

\section{Related Work}
\label{sec:related}
\paragraph{Visual Hallucination}
The study of visual hallucination has been shaped largely by benchmark design. POPE~\citep{li2023pope} cast the problem as binary object-existence classification; CHAIR~\citep{rohrbach2018chair} measures hallucinated objects in free-form captions; AMBER~\citep{wang2023amber} and MMHal-Bench~\citep{sun2023aligning} extend evaluation to attribute and relational errors; and HallusionBench~\citep{liu2023hallusionbench} targets counter-commonsense scenarios (see \citet{chen2025survey_multimodal_hallucination} for a comprehensive survey). On the mitigation side, approaches span training-time alignment~\citep{yu2024rlhfv,yu2023hallucidoctor}, decoding-time contrast~\citep{huang2024opera,leng2024vcd}, and post-hoc correction~\citep{yin2023woodpecker}. All of these treat the model as a black box. Our work is complementary: we characterize the internal circuits that produce hallucination, providing a mechanistic foundation for future mitigation.

\paragraph{Activation Patching} Activation patching~\citep{meng2022locating} has become a standard tool for circuit discovery in language models, with applications to indirect object identification~\citep{wang2023interpretability}, automated subgraph search~\citep{conmy2023automated}, and edge-level information flow~\citep{goldowskydill2023localizing}. Recent work extends these techniques to VLMs: \citet{neo2025towards_interpreting_vlm} applied logit lens and activation patching to visual information processing; \citet{qli2025causal_tracing_vlm} proposed cross-modal causal tracing with an inference-time intervention; and \citet{rudman2026mechanisms_pih} identified attention heads responsible for prompt-induced hallucination in three VLMs. These studies operate at the level of individual components. Our analysis instead characterizes pathway-level interaction structure across architecturally diverse VLMs, which surfaces the cross-subset polarity flip in grounding components that single-component analysis cannot see.

\paragraph{Hallucination Mitigation} A separate line of work modifies internal activations at inference time to reduce hallucination, including ITI~\citep{li2024inference}, VISTA~\citep{li2025vista}, DMAS~\citep{yin2026dmas}, and SteerVLM~\citep{sivakumar2025steervlm}~\citep[see also][]{kogilathota2026halp,su2025asd,turner2024activation}. These methods target mitigation and make varying assumptions about the geometry of the hallucination signal, from a single static linear direction to per-input or learned ones. Our work is not a mitigation contribution; we use ITI and mean-difference projection in Section~\ref{sec:intervention_results} as a controlled probe of the static-direction assumption on the same components our circuit identifies.

\section{Methodology}
\label{sec:method}

\begin{figure}[t]
\centering
\includegraphics[width=\linewidth]{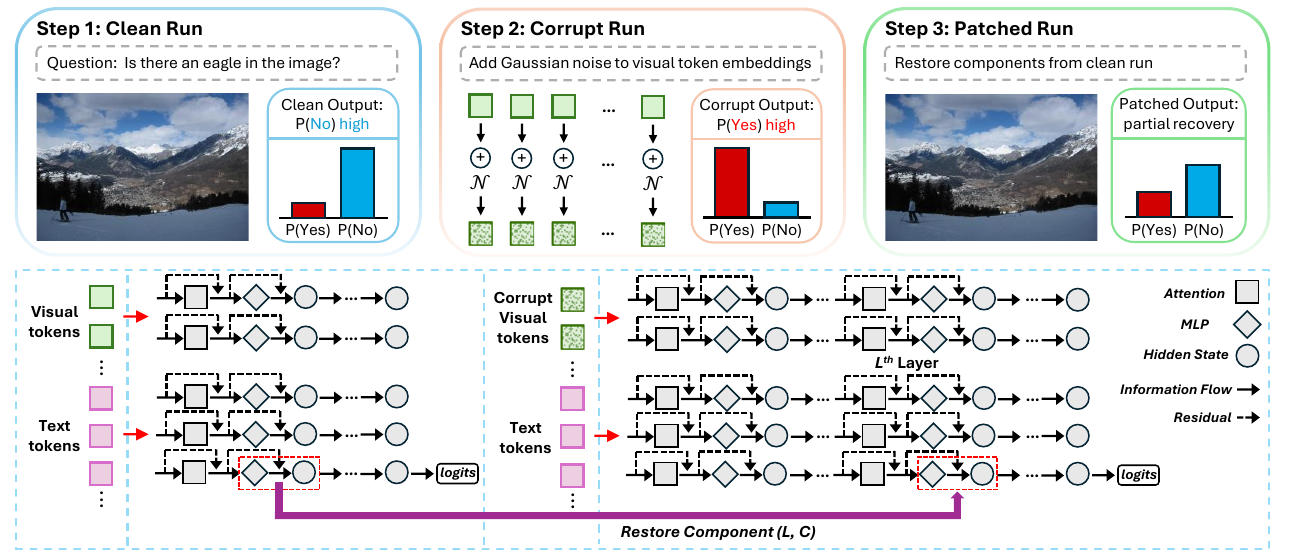}
\caption{
Overview of the activation-patching pipeline for identifying hallucination-related components in VLMs.
For each image--question pair, we first run a clean pass, then corrupt visual token embeddings with Gaussian noise, and finally restore selected components from the clean run into the corrupt run.
The recovery of the correct logit difference after restoration estimates the causal contribution of each component, which is later used to extract grounding and hallucination pathways.
}
\label{fig:overview}
\end{figure}

\subsection{Activation Patching for Vision-Language Models}
\label{sec:patching}

Activation patching~\citep{meng2022locating} was developed for text-only models, where corruption is applied to token embeddings. Extending it to VLMs is non-trivial: visual information enters the network through architecture-specific mechanisms, including direct embedding concatenation (Qwen3-VL), learned MLP projectors (LLaVA, InternVL3), and dedicated cross-attention layers (Llama-3.2-Vision), each requiring a different patching interface. We therefore implement a unified patching framework with model-specific adapters for the four VLM families in our study, and corrupt visual rather than text tokens. Appendix A gives the full protocol.

For each image/question pair we run a clean pass, a corrupt pass (Gaussian noise at $3\sigma$ on all visual tokens at layer~0, sufficient to drive accuracy to chance), and a patch pass that restores a single component $(L, C)$ to its clean value. The \textbf{indirect effect} of $(L, C)$ is
\begin{equation}
\mathrm{IE}(L, C) = \Delta_{\text{patched}}(L, C) - \Delta_{\text{corrupt}},
\label{eq:ie}
\end{equation}
where $\Delta = \text{logit}(\text{correct token}) - \text{logit}(\text{incorrect token})$ at the final token position. A large positive IE indicates the component is a critical conduit for visual information. The \textbf{total effect} $\mathrm{TE} = \Delta_{\text{clean}} - \Delta_{\text{corrupt}}$ normalizes IE across samples.

\subsection{Circuit Extraction}
\label{sec:circuit_extraction}

We run the patching procedure on $n{=}1{,}000$ POPE-adversarial samples per model. For each sample, the model either correctly answers “no” to a non-existent object or hallucinates by incorrectly answering “yes.” We partition the samples accordingly and, for each component, compare the normalized IE distributions between the two groups using Welch's $t$-test. All $p$-values are corrected for multiple comparisons via the Benjamini-Hochberg procedure at a false discovery rate of $\alpha{=}0.05$. A component is included in the hallucination circuit if it satisfies two criteria: $p < 0.05$ after correction, and $|\text{Cohen's } d| > 0.3$.

The sign of Cohen's $d$ determines the pathway assignment. Components with $d > 0$, meaning their indirect effect is larger when the model answers correctly, belong to the \textbf{visual grounding pathway}. Components with $d < 0$, meaning their indirect effect is larger during hallucination, belong to the \textbf{hallucination pathway}.

\subsection{Pathway-Level Interaction Decomposition}
\label{sec:cpa}

Single-component patching identifies which components matter but leaves open whether a group of components carries overlapping information, cooperates, or contributes independently. We address this at the pathway level with a diagnostic we call Conditional Pathway Analysis (CPA).

If multiple components each carry the same evidence, removing any one of them still leaves the signal intact in the others, so the joint effect of restoring them all is smaller in magnitude than the sum of their individual effects (redundancy). If instead each component carries only a fragment and only their combination is sufficient to drive the prediction, the joint effect exceeds the sum in magnitude (synergy). The relationship between joint and summed effects thus indicates how the pathway is processing information.

We compute interaction within each pathway separately rather than over the full set of significant components, because grounding and hallucination components have opposite-signed causal effects, and a single joint-minus-sum term over the whole circuit would mix interactions of opposite sign.

For a pathway $\mathcal{P} = \{C_1, \ldots, C_k\}$, the \textbf{joint effect} $\mathrm{IE}(\mathcal{P})$ is the indirect effect obtained when all pathway components are restored simultaneously. The interaction term subtracts the sum of individual effects:
\begin{align}
\mathrm{IE}(\mathcal{P}) &= \Delta_{\text{patched}(\mathcal{P})} - \Delta_{\text{corrupt}}, \label{eq:joint} \\
\mathrm{Interaction}(\mathcal{P}) &= \mathrm{IE}(\mathcal{P}) - \sum_{i=1}^{k} \mathrm{IE}(C_i). \label{eq:interaction}
\end{align}
Because the polarity of $\mathrm{IE}(C_i)$ may differ between correct and hallucinating subsets, the sign of Eq.~\ref{eq:interaction} alone conflates redundancy with the underlying IE polarity. We therefore additionally report a polarity-free magnitude metric:
\begin{equation}
\mathrm{MagDiff}(\mathcal{P}) = \bigl|\mathrm{IE}(\mathcal{P})\bigr| - \sum_{i=1}^{k} \bigl|\mathrm{IE}(C_i)\bigr|, \label{eq:magdiff}
\end{equation}
where negative values uniformly indicate magnitude-redundancy and positive values indicate magnitude-synergy. We use $\mathrm{MagDiff}$ as the primary diagnostic for within-subset interaction structure and report the per-subset mean of $\mathrm{IE}(C_i)$ separately to expose any polarity flip across subsets. We compute these per sample on $n{=}400$ POPE-adversarial samples per model, and test (i) whether $\mathrm{MagDiff}$ differs from zero (one-sample $t$-test) and (ii) whether the mean $\mathrm{IE}(C_i)$ shifts between correct and hallucinating samples (two-sample $t$-test).

\subsection{Logit Lens Validation}
\label{sec:logitlens_method}

To validate the pathway labels independently of the patching procedure, we ask whether layers assigned to the grounding vs.\ hallucination pathway by our patching-based criterion have differential layer-wise contributions to the correct answer. We answer this using logit lens analysis~\citep{nostalgebraist2020logitlens} as a decoding tool. At each layer $l$, we project the hidden state through the final layer norm and language model head to obtain intermediate logits: $\hat{y}_l = \text{lm\_head}(\text{norm}(h_l))$. We compute the per-layer delta, the change in $\Delta$ relative to the preceding layer, which reveals each layer's marginal contribution to the final prediction. If the pathway labels are functionally accurate, grounding-pathway layers should have positive delta and hallucination-pathway layers negative delta on correct samples, with polarities reversing on hallucinating samples.

\subsection{Cross-Architecture Comparison}
\label{sec:crossarch_method}

Because the five models have different numbers of layers (28--48), direct layer-by-layer comparison is not meaningful. We normalize all layer indices to a relative depth in $[0, 1]$ and compare at two levels of granularity. At the macro level, we test whether the spatial distribution of pathway components is shared across models, using Fisher's exact test on $2 \times 2$ contingency tables. At the micro level, we compute Pearson correlations on depth-normalized effect-size profiles and apply TOST equivalence testing with a bound of $|r| < 0.3$.

\subsection{Targeted Intervention}
\label{sec:intervention_method}

For models whose hallucination pathway is sufficiently concentrated, we suppress the identified components by scaling their outputs during inference:
\begin{equation}
h'_{\text{out}} = s \cdot h_{\text{out}}, \quad s \in \{0.0, 0.25, 0.5, 0.75\}.
\label{eq:scaling}
\end{equation}
This is applied at every forward pass to the output of each hallucination-pathway component ($d < 0$ in the circuit extraction). No retraining or additional data is required. We select the optimal $s$ per model on a held-out selection set (see Section~\ref{sec:setup}) and evaluate the selected configuration on a disjoint evaluation set.

For models with dispersed circuits (many hallucination-pathway components), we additionally introduce \textbf{top-$k$ component selection}: we rank hallucination-pathway components by $|d|$ and suppress only the top-$k$ strongest, evaluating $k \in \{3, 5, 8, 10, 15, 20, 30\}$. The optimal $k$ is selected on the same selection set used for $s$, and reported on the disjoint evaluation set.

We compare against two linear-steering baselines from prior work~\citep{li2024inference,zou2023representation}, applied to the same hallucination-pathway components selected by our circuit. \textbf{Mean-difference projection} projects out $\hat{d} = \text{normalize}(\mu_{\text{halluc}} - \mu_{\text{correct}})$ from each component's activations: $h' = h - \alpha (h \cdot \hat{d}) \hat{d}$. \textbf{Probe-based ITI}~\citep{li2024inference} replaces $\hat{d}$ with a logistic-regression probe direction trained per attention head and adds it to the top-$K$ heads. Both baselines use symmetric $\alpha$ grids; including negative $\alpha$ is essential because the probe direction points toward the hallucination class (Appendix~\ref{app:iti}), so reducing hallucination requires subtracting it. As a further control, we include a \textbf{random control} that applies the uniform-scaling factor to the same number of randomly selected components. Full grids, probe-training details, and tie-breaking rules are in Appendix~\ref{app:directional}.

\section{Experimental Setup}
\label{sec:setup}

\paragraph{Models.}
We select five VLMs that together cover three LLM backbone families, four vision--language integration strategies (embed-concat, projector-concat, ViT-MLP-LLM, cross-attention), and two model scales (Table~\ref{tab:setup}). InternVL3-8B and InternVL3-14B share an architecture but use different LLM backbones, letting us test scale sensitivity within the same family. HuggingFace identifiers are listed in Appendix~\ref{app:implementation}.

\paragraph{Data and evaluation.}
All patching experiments use the adversarial split of POPE~\citep{li2023pope}, which selects negative-sample objects based on co-occurrence frequency. We use 1{,}000 samples for circuit discovery, 400 for CPA, 200 for logit lens analysis, and 500 for intervention evaluation. These four sets are drawn independently from POPE-adversarial; the patching set and the intervention set are disjoint. Of the 500 intervention samples, 100 are reserved as a selection set for choosing $s$ and $k$; the remaining 400 form the held-out evaluation set on which final numbers in Sections~\ref{sec:intervention_results} and~\ref{sec:cross_benchmark} are computed. For cross-benchmark evaluation, we test on the popular and random POPE splits and on AMBER~\citep{wang2023amber}, which covers existence, attribute, and relational hallucination types. Hyperparameters are selected only on the POPE-adversarial selection set and applied unchanged to all other benchmarks, making the cross-benchmark evaluations pure transfer tests.

\paragraph{Patching configuration.}
We corrupt visual embeddings with Gaussian noise at $3\times$ the standard deviation of clean visual tokens. All patching operates at the component level (one attention sublayer and one MLP sublayer per transformer block). All evaluations use greedy decoding to ensure reproducibility.

\begin{table}[t]
\centering
\caption{\textbf{Models and experimental setup.} All patching experiments use POPE-adversarial ($n{=}1{,}000$). Corruption: Gaussian noise ($3\sigma$) on visual token embeddings at layer~0. Llama-3.2-Vision uses 8 cross-attention layers (of 40 total) for visual input.}
\label{tab:setup}
\begin{adjustbox}{max width=\linewidth}
\begin{tabular}{lccccccc}
\toprule
\textbf{Model} & \textbf{Arch.\ type} & \textbf{LLM backbone} & \textbf{Layers} & \textbf{Sig.} & \textbf{Grnd.} & \textbf{Hall.} & \textbf{Halluc.\ rate} \\
\midrule
Qwen3-VL-8B   & Embed-concat       & Qwen3 (8B)     & 36 & 26 & 14 & 12 & 13.1\% \\
LLaVA-v1.6-7B & Projector-concat   & Mistral (7B)   & 32 & 48 & 18 & 30 & 13.4\% \\
Llama-3.2-V-11B & Cross-attention  & Llama-3 (11B)  & 40 & 31 & 11 & 20 & 15.9\% \\
InternVL3-8B  & ViT-MLP-LLM        & Qwen2.5 (8B)   & 28 & 27 & 21 & 6  & 12.6\% \\
InternVL3-14B & ViT-MLP-LLM        & Qwen2.5 (14B)  & 48 & 45 & 29 & 16 & 13.4\% \\
\bottomrule
\end{tabular}
\end{adjustbox}
\end{table}

\section{Experiments}
\label{sec:results}

\subsection{A Dual-Pathway Circuit for Visual Hallucination}
\label{sec:circuits}

Activation patching reveals a consistent dual-pathway organization across all studied models (Figure~\ref{fig:dual_pathway}). The strongest components per model are summarized in Appendix Table~\ref{tab:top_components}. In each model, the set of significant components ($p < 0.05$ after BH correction, $|d| > 0.3$) splits into two groups that differ in sign, depth profile, and functional role.

The \textbf{visual grounding pathway} ($d > 0$) consists of components whose indirect effect is substantially higher on correct samples, with peak effect sizes reaching $d = +1.59$. The \textbf{hallucination pathway} ($d < 0$) is preferentially active during hallucination, with peak $|d|$ up to $0.89$. In Llama-3.2-Vision, the strongest grounding component is one of the model's dedicated cross-attention layers (L3.attn, $d = +0.91$), providing an architectural sanity check on the patching-based classification. LLaVA exhibits the most dispersed hallucination pathway, with 30 of its 48 significant components having $d < 0$ and spanning the entire network depth.

Circuit sizes range from 26 to 48 significant components per model, with hallucination-pathway share varying from 22\% (InternVL3-8B) to 65\% (Llama-3.2). The dual-pathway organization is preserved across $|d|$ thresholds in $\{0.25, 0.30, 0.35\}$ for all five models, with both pathways always retaining at least 5 components (Appendix~\ref{app:robustness}).

\begin{figure*}[t]
\centering
\includegraphics[width=\linewidth]{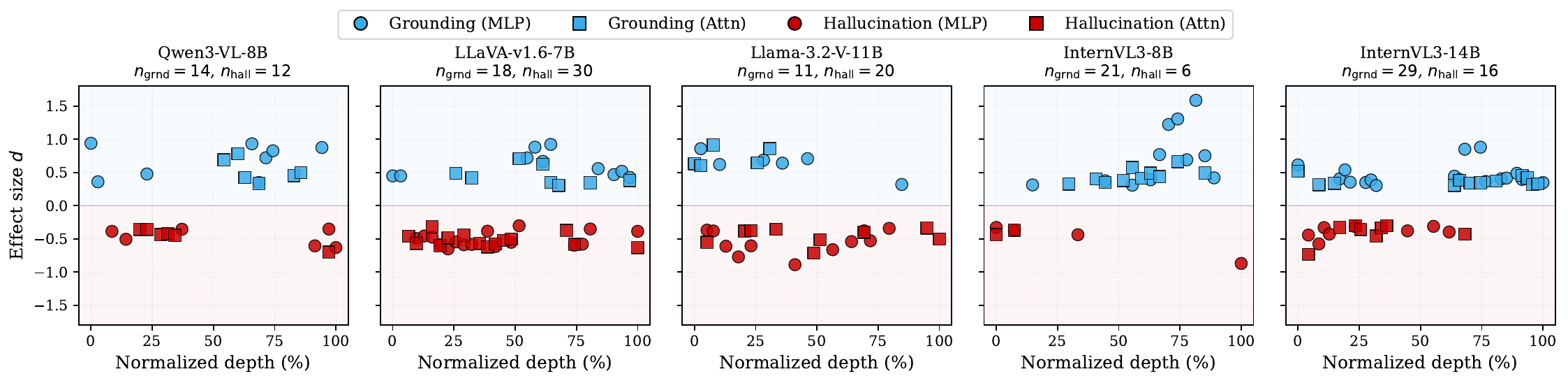}
\caption{\textbf{Dual-pathway organization across all five VLMs.} Hallucination components (red) concentrate at early layers and at network boundaries; grounding components (blue) concentrate at mid-to-late depths. Each panel shows components satisfying the inclusion criteria ($p_{\text{adj}} < 0.05$, $|d| > 0.3$; $n=1{,}000$) plotted by normalized depth ($x$-axis) and Cohen's $d$ ($y$-axis). See Appendix Table~\ref{tab:top_components} for the strongest components per model.}
\label{fig:dual_pathway}
\end{figure*}

\subsection{CPA Reveals Consistent Redundancy and a Grounding-Pathway Polarity Flip}
\label{sec:cpa_results}

\textbf{Both pathways are redundant within each subset; the grounding pathway additionally exhibits a consistent IE polarity flip between subsets.} The polarity-free magnitude diagnostic $\mathrm{MagDiff}$ shows that, in every model and every subset, the joint pathway effect is substantially smaller in magnitude than the sum of individual effects: across the $5{\times}2{\times}2 = 20$ (model, pathway, subset) cells, the magnitude ratio $|\mathrm{IE}(\mathcal{P})|/\sum_i|\mathrm{IE}(C_i)|$ stays in $[0.07, 0.69]$ with a mean of $0.32$, indicating strong within-subset redundancy throughout (Figure~\ref{fig:cpa}; full numerical values in Appendix Table~\ref{tab:cpa_full}). The shift in $\mathrm{MagDiff}$ between subsets is small and inconsistent in sign across models for both pathways (grounding $|d|$ from $0.13$ to $0.92$; hallucination $|d|$ from $0.14$ to $0.59$), giving no support for a redundancy-to-synergy transition under the polarity-free metric.

What does differ systematically between subsets, and only for the grounding pathway, is the polarity of the individual indirect effects. On correct samples, $61$--$83\%$ of grounding-pathway component IEs are positive, with subset-mean IEs of $+0.13$ to $+1.59$ across the five models. On hallucinating samples, the same components flip: only $21$--$33\%$ of IEs remain positive, and the subset-mean IE turns negative ($-1.08$ to $-0.07$) in every model (cross-subset Cohen's $d$ from $-1.60$ to $-2.45$, all $p < 0.001$). The hallucination pathway shows no such consistent flip: the mean-IE shift is smaller in magnitude (mean $|d|=0.59$ vs.\ $1.92$ for grounding) and reverses direction across models, with InternVL3-8B showing no flip at all (mean IE remains negative in both subsets, $-0.09$ and $-0.16$).

The mechanistic reading is that on hallucinating samples, restoring a grounding-pathway component pulls the corrupt run toward the model's clean-run prediction, which on these samples is the hallucinated answer. The within-subset redundancy persists in both phases rather than dissolving when the model hallucinates, suggesting that the language-prior bias does not break apart the redundant copies carried by the grounding pathway but redirects what they encode. The failure on these samples is thus not that visual evidence is missing or weak, but that the ``grounding'' representations themselves have been entrained to the model's non-visual output.

\begin{figure}[t]
\centering
\includegraphics[width=0.75\linewidth]{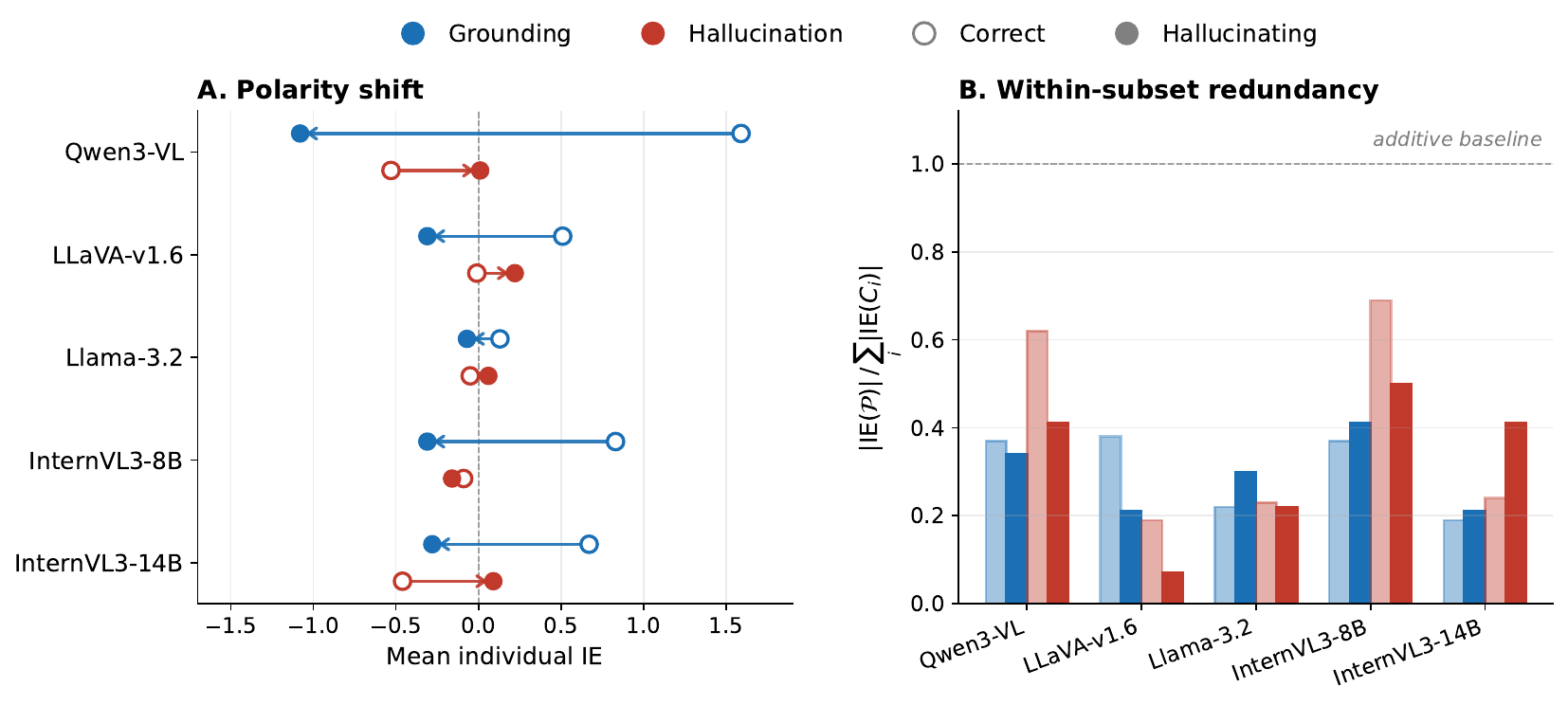}
\caption{\textbf{Pathway-level CPA diagnostics across all five VLMs.} \textbf{Left:} mean individual indirect effect per pathway and subset; arrows point from correct (hollow) to hallucinating (filled) samples. The grounding pathway exhibits a consistent polarity flip from positive on correct samples to negative on hallucinating samples in every model; the hallucination pathway shows no consistent flip. \textbf{Right:} magnitude ratio $|\mathrm{IE}(\mathcal{P})|/\sum_i |\mathrm{IE}(C_i)|$, with light bars = correct and dark bars = hallucinating samples. All cells fall well below $1.0$ ($0.07$--$0.69$, mean $0.32$), indicating strong within-subset redundancy regardless of subset. Numerical values are in Appendix Table~\ref{tab:cpa_full}.}
\label{fig:cpa}
\end{figure}

\subsection{Logit Lens Supports the Functional Roles}
\label{sec:logitlens_results}

Logit lens analysis provides patching-independent evidence that the pathway labels track functional differences (full per-model values in Appendix~\ref{app:logitlens}, Table~\ref{tab:logitlens}). If grounding-pathway layers drive image-faithful predictions and hallucination-pathway layers push toward language-prior errors, their layer-wise contributions should shift direction between correct and hallucinating samples. In four of five models (Qwen3-VL, LLaVA, InternVL3-8B, InternVL3-14B), the differential is significant and in the predicted direction, with $d \in [-1.25, -0.65]$ (all $p < 0.05$).

Llama-3.2-Vision is the exception: its hallucinating-sample differential is not significant ($d = -0.18$, $p = 0.43$). Llama-3.2 injects visual information through dedicated cross-attention layers interleaved with self-attention, so grounding- and hallucination-pathway layers do not form a single layer-wise logit trajectory of the kind logit lens decodes. The pathway labels are nonetheless functionally validated for this model by its strongest hallucination-pathway concentration in our circuits (Table~\ref{tab:setup}) and its cleanest response to the targeted intervention (Section~\ref{sec:intervention_results}).

\subsection{Cross-Architecture Analysis: Shared Organization, Distinct Wiring}
\label{sec:crossarch}

We compare circuits across architectures along two axes, normalizing layer indices to a relative depth in $[0, 1]$ to handle differing network depths (full pairwise statistics in Appendix~\ref{app:crossarch}). At the macro level, hallucination-pathway components concentrate in early layers and at network boundaries while grounding-pathway components concentrate in mid-to-late layers, and this spatial pattern is consistent across models (Fisher's exact test, all six pairs $p > 0.05$ against a shared-pattern null). At the micro level, layer-by-layer profile correlations are uniformly low ($|r| \leq 0.33$), and the within-family InternVL3-8B/14B comparison is no more similar than cross-family pairs. The combination is most consistent with the dual-pathway organization arising during VLM training rather than being inherited from a specific backbone or architecture.

\subsection{Causal Validation: Suppressing Hallucination-Pathway Components}
\label{sec:intervention_results}

\begin{figure}[t]
\centering
\includegraphics[width=0.7\linewidth]{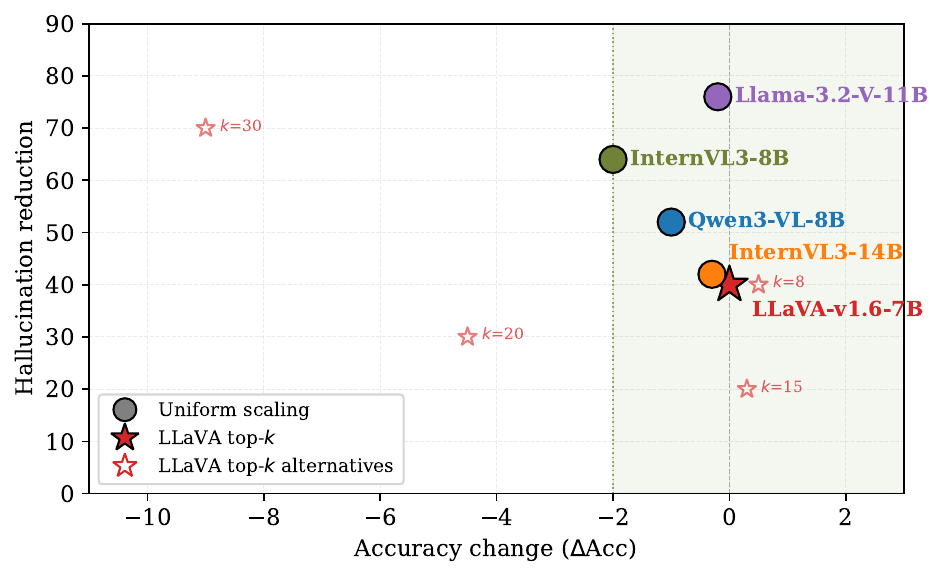}
\caption{\textbf{Intervention Pareto front on POPE-adversarial ($n{=}400$ held-out).} Filled markers show each model's selected configuration; $x$-axis: accuracy change (pp), $y$-axis: relative hallucination reduction. All five models achieve 40--76\% reduction at $\leq$2\,pp accuracy cost. LLaVA uses top-$k{=}10$ at $s{=}0.5$; hollow stars show alternative $k$ values. Full numerical values are in Appendix Table~\ref{tab:intervention_full}.}
\label{fig:pareto}
\end{figure}

Figure~\ref{fig:pareto} shows that uniform scaling of hallucination-pathway outputs substantially reduces hallucination in every model; full numerical values are in Appendix Table~\ref{tab:intervention_full}. All numbers reported in this section are computed on the held-out 400-sample test set, with the optimal $s$ selected on the disjoint 100-sample selection set (setup in Section~\ref{sec:setup}). Qwen3-VL responds best to $s{=}0.25$, while Llama-3.2, InternVL3-8B, and InternVL3-14B respond best to $s{=}0.5$. Relative hallucination reductions on POPE-adversarial range from 40\% (LLaVA-v1.6) to 76\% (Llama-3.2) at $\leq$2\,pp accuracy cost.

\paragraph{Top-$k$ ablation resolves LLaVA's dispersed circuit.}
LLaVA's hallucination pathway spans 30 components. Suppressing all of them ($s{=}0.5$) drops accuracy by 9.0\,pp. We therefore evaluate suppressing only the $k$ strongest components for $k\in\{3,5,8,10,15,20,30\}$, selecting $k$ on the 100-sample selection set. The selected configuration is $k{=}10$, which yields a 40\% relative hallucination reduction (5.0\% $\to$ 3.0\%) at no accuracy cost (Figure~\ref{fig:pareto}). Top-$k$ selection thus identifies a compact core of strongly hallucination-driving components within an otherwise dispersed circuit, extending the intervention to models where the $|d| > 0.3$ threshold alone is too permissive.

\paragraph{Static-direction interventions are partially effective but bounded.}
To probe whether the hallucination pathway is well described by a single static linear direction, we apply mean-difference projection and probe-based ITI~\citep{li2024inference} to the same hallucination-pathway components, with hyperparameters selected under the same protocol as uniform scaling (full grids, per-model results, and probe-quality checks in Appendix~\ref{app:directional}). ITI reduces hallucination on every evaluated model but consistently less than per-component scaling, while mean-difference projection fails to yield any reduction on the held-out test set across the three models evaluated. The per-model relative effect tracks the per-component activation geometry: the leading singular direction of the per-component hallucination signal captures only 45--69\% of its variance (Appendix~\ref{app:geometry}), bounding what any single static direction can do. Per-component scaling sidesteps this bound by treating the pathway as a set of adjustable contributions rather than a single direction.

A random-component control at the same scaling factor produces no systematic reduction (mean $|\Delta\text{H}| = 3.0$\,pp with the sign positive on all three evaluated models, indicating that random suppression tends to slightly increase rather than decrease hallucination; the largest such increase is $+7.2$\,pp on Llama-3.2), confirming that the effect depends on which components are scaled, not on scaling alone. The probes are discriminative (per-model mean head-probe validation accuracy 0.62--0.67, hundreds of heads above 0.75), ruling out probe-fitting failure as the cause of ITI's bounded effect. As a specificity check, applying the same uniform-scaling protocol to the grounding pathway at $s{=}0$ collapses POPE accuracy to chance ($0.50$--$0.51$) on every model (Appendix~\ref{app:grounding_suppress}), establishing a double dissociation: hallucination-pathway suppression reduces hallucination at near-zero accuracy cost, while grounding-pathway ablation destroys task performance.

\subsection{Cross-Benchmark Generalization}
\label{sec:cross_benchmark}

To test whether the hallucination pathways identified on POPE-adversarial generalize beyond object-existence questions, we evaluate the same targeted intervention on two additional settings: all three POPE splits and the AMBER benchmark~\citep{wang2023amber}, which covers three hallucination types. Circuits and hyperparameters are identical to those selected in Section~\ref{sec:intervention_results}. No tuning is done on these new benchmarks, and all of them are pure generalization evaluations.

\paragraph{POPE splits.}
The intervention generalizes consistently across POPE difficulty levels (Appendix Table~\ref{tab:cross_benchmark_full}). Hallucination reduction is strongest on the adversarial split, where baseline hallucination is highest, and attenuates on the random split, where baseline hallucination is already low, consistent with a floor effect. Accuracy cost remains within 2\,pp for most models.

\paragraph{AMBER: type-dependent transfer.}
The intervention transfers to AMBER with a clear type-dependent pattern. On existence questions, the type semantically closest to POPE, four of five models show improvement (10--42\% relative reduction). On relational hallucination, Qwen3-VL, InternVL3-8B, and Llama-3.2 achieve 36--47\% relative reduction, suggesting substantial overlap between the object-existence and relational hallucination circuits. On attribute questions, only Llama-3.2 ($-$23\%) and InternVL3-8B ($-$17\%) show meaningful improvement; the remaining models are unaffected. This asymmetry suggests that attribute hallucination involves partially distinct circuits not captured by the POPE-derived pathway.

\paragraph{Model-specific patterns.}
Llama-3.2-Vision shows the strongest cross-benchmark generalization, with hallucination reduction across all six evaluation settings, consistent with its cross-attention architecture producing a more concentrated hallucination pathway. InternVL3-14B shows the opposite profile: strong POPE reductions (29--68\%) but weak AMBER transfer; we discuss possible mechanistic correlates (the largest cross-subset polarity-flip magnitude, Section~\ref{sec:cpa_results}) in Appendix~\ref{app:crossarch}.

\section{Discussion}
\label{sec:discussion}

\paragraph{What generalizes, and implications for intervention design.}
The macro-level dual-pathway organization recurs despite differences in backbone, fusion strategy, and scale, while micro-level wiring differs even within the same backbone family (Section~\ref{sec:crossarch}), pointing to training rather than architecture as the origin of the organization. Across hallucination types, the intervention transfers well to relational hallucination in three of five models but leaves attribute hallucination largely unaffected. One interpretation is that relational hallucination, like object-existence hallucination, is driven primarily by language priors such as co-occurrence statistics, whereas attribute hallucination may involve a different failure mode such as imprecise visual feature extraction. The pathway asymmetry also constrains intervention design. The hallucination pathway does not show a consistent polarity flip, static-direction baselines are bounded by per-component activation geometry, and hallucination-signal variance is spread across many singular directions (Appendix~\ref{app:geometry}). These results suggest that no single static direction captures the full pathway contribution. Per-component scaling sidesteps this bound for our causal test, but does not constitute a deployable mitigation method; designing one under this constraint is a separate engineering problem.

\paragraph{Conclusion.}
We identified a shared dual-pathway organization for object-existence hallucination across five architecturally diverse VLMs, and showed that pathway-level analysis exposes a consistent IE polarity flip in grounding components between correct and hallucinating samples, a signature invisible to single-component patching. Causal validation by component scaling, together with the bounded effect of matched static-direction interventions, indicates that the hallucination pathway acts through multiple distinct activation directions rather than a single shared one. The selective transfer to relational but not attribute hallucination points to partially distinct circuits underlying different hallucination types; direct circuit extraction on attribute- and relation-focused benchmarks is a natural next step. We release our code  to support further work.


\bibliographystyle{plainnat}
\bibliography{references}


\appendix

\section{Implementation Details}
\label{app:implementation}

\paragraph{Models and HuggingFace identifiers.}
The five VLMs evaluated are: Qwen3-VL-8B~\citep{qwen3vl2025} (\texttt{Qwen/Qwen3-VL-8B-Instruct}), which concatenates visual tokens directly into the embedding sequence; LLaVA-v1.6-7B~\citep{liu2023llava} (\texttt{llava-hf/llava-v1.6-mistral-7b-hf}), which uses a learned MLP projector into the Mistral-7B token space; Llama-3.2-Vision-11B~\citep{meta2024llama32} (\texttt{meta-llama/Llama-3.2-11B-Vision-Instruct}), which injects visual information through dedicated cross-attention layers interleaved with standard self-attention; and InternVL3-8B / 14B~\citep{zhu2025internvl3} (\texttt{OpenGVLab/InternVL3-8B-hf}, \texttt{OpenGVLab/InternVL3-14B-hf}), which route vision-encoder outputs through an MLP bridge into Qwen2.5-8B / 14B backbones.

\paragraph{Activation patching protocol.}
Our patching implementation adapts the procedure of \citet{meng2022locating} to handle the diversity of VLM architectures. We implement model-specific adapters for each VLM family (Qwen, LLaVA, InternVL, Llama) that handle differences in visual token placement, attention layer structure (self-attention vs.\ cross-attention), and embedding concatenation strategy. The core patching loop is architecture-agnostic: it operates on cached activations and patched forward passes through the language-model backbone.

For the corrupt run, we add Gaussian noise with standard deviation $3 \times \sigma_{\text{visual}}$ to all visual token positions at layer~0, where $\sigma_{\text{visual}}$ is the empirical standard deviation of the clean visual token embeddings computed per sample. We verified that this noise level is sufficient to destroy visual information: the accuracy of the corrupt run drops to near chance level (50\%) for all five models.

\paragraph{CPA computation.}
The joint patching required for CPA is implemented by simultaneously restoring all pathway components in a single patch run. For a pathway with $k$ components, this requires one additional forward pass per sample (beyond the $2k$ passes for individual component patching), making CPA computationally efficient. We run CPA on 400 POPE-adversarial samples per model.

\paragraph{Logit lens.}
At each layer, we apply the final layer norm and language model head to the hidden state at the last token position (the position where the model generates ``Yes'' or ``No''). We record $P(\text{Yes})$, $P(\text{No})$, and the logit difference at each layer. The per-layer delta is computed as $\Delta_l = \Delta(\text{logit\_diff})_l - \Delta(\text{logit\_diff})_{l-1}$. We use 200 POPE-adversarial samples per model.

\paragraph{Hardware and runtime.}
All experiments were conducted on NVIDIA H200 144GB GPUs. Component-level patching on 1{,}000 samples takes approximately 2--18 hours per model depending on model size and sequence length (Llama-3.2 is fastest at $\sim$1.8s/sample; InternVL3-14B is slowest at $\sim$18s/sample). CPA adds roughly 2--3 hours per model for 400 samples. Logit lens analysis on 200 samples completes in under 1 hour per model. The full pipeline for one model runs in under 24 hours on a single GPU.

\section{Threshold Sensitivity for Circuit Extraction}
\label{app:robustness}

The circuit-extraction criterion in Section~\ref{sec:circuit_extraction} uses two thresholds: BH-corrected $p < 0.05$ and $|\text{Cohen's } d| > 0.3$. The $p$-value cutoff follows standard FDR practice. To evaluate sensitivity to the effect-size cutoff, we re-extract circuits at $|d| \in \{0.25, 0.30, 0.35\}$ from the same patching results (no additional model runs required), and measure overlap with the reference $|d| > 0.3$ circuit using Jaccard similarity computed separately for each pathway.

Results are reported in Table~\ref{tab:d_robustness}. The dual-pathway organization is preserved at every threshold for all five models: each model retains at least 5 grounding-pathway and 5 hallucination-pathway components throughout the $\pm 0.05$ window, and 16 of the 20 (model, pathway, perturbation) cells have Jaccard $\geq 0.80$ (minimum $0.60$). The qualitative finding, that both pathways exist and that the grounding pathway dominates in size, is preserved at every threshold.

\begin{table}[!htbp]
\centering
\small
\caption{\textbf{Circuit composition across $|d|$ thresholds.} For each model, we report the number of grounding-pathway and hallucination-pathway components at three thresholds, plus Jaccard similarity of pathway membership relative to the paper's $|d| > 0.3$ circuit. Computed post-hoc on the same patching results used in the main paper; no additional GPU runs were performed.}
\label{tab:d_robustness}
\begin{tabular}{lccccccc}
\toprule
& \multicolumn{2}{c}{$|d| > 0.25$} & \multicolumn{2}{c}{$|d| > 0.30$ (paper)} & \multicolumn{2}{c}{$|d| > 0.35$} \\
\cmidrule(lr){2-3} \cmidrule(lr){4-5} \cmidrule(lr){6-7}
\textbf{Model} & $n_{\text{grnd}}/n_{\text{hall}}$ & $J_{\text{grnd}}/J_{\text{hall}}$ & $n_{\text{grnd}}/n_{\text{hall}}$ & $J$ (ref) & $n_{\text{grnd}}/n_{\text{hall}}$ & $J_{\text{grnd}}/J_{\text{hall}}$ \\
\midrule
Qwen3-VL-8B    & 20/16 & 0.70/0.75 & 14/12 & 1.00 & 12/12 & 0.86/1.00 \\
LLaVA-v1.6-7B  & 20/32 & 0.90/0.94 & 18/30 & 1.00 & 15/28 & 0.83/0.93 \\
Llama-3.2-V-11B & 15/22 & 0.73/0.91 & 11/20 & 1.00 & 10/18 & 0.91/0.90 \\
InternVL3-8B   & 22/10 & 0.96/0.60 & 21/6  & 1.00 & 18/5  & 0.86/0.83 \\
InternVL3-14B  & 32/20 & 0.91/0.80 & 29/16 & 1.00 & 19/9  & 0.66/0.56 \\
\bottomrule
\end{tabular}
\end{table}

\section{Statistical Testing Details}
\label{app:stats}

For CPA (Section~\ref{sec:cpa}), we perform two planned tests per pathway per model (one-sample $t$-test and correct-vs-hallucinating two-sample $t$-test), yielding 4 tests per model or 20 across all five models. For logit lens (Section~\ref{sec:logitlens_method}), we perform one differential test per model (5 tests). Because these are a small number of pre-specified, hypothesis-driven comparisons rather than exploratory screening, we report uncorrected $p$-values but note that all key findings (grounding-pathway differential $d$, all $p < 0.001$) survive Bonferroni correction at any conventional $\alpha$ level. BH-corrected and TOST equivalence-test details for circuit extraction and cross-architecture comparison are in Sections~\ref{sec:circuit_extraction} and~\ref{sec:crossarch}; full pairwise TOST values are in Table~\ref{tab:crossarch}.

\section{Linear-Steering Baselines: Full Numerical Results}
\label{app:directional}

This appendix gives the complete numerical results underlying Section~\ref{sec:intervention_results} for both linear-steering baselines: basic directional suppression (mean-difference projection) and probe-based ITI~\citep{li2024inference}.

\subsection{Held-Out Test-Set Comparison}
\label{app:baseline_comparison}

Table~\ref{tab:linear_steering_baselines} reports the held-out test-set comparison for all four interventions side by side. Directional suppression projects out the per-component mean-difference direction $h' = h - \alpha (h \cdot \hat{d})\hat{d}$ at $\alpha \in \{\pm 0.5, \pm 1.0, \pm 1.5\}$, with $\alpha$ selected on the same selection set and accuracy budget as uniform scaling. Directional suppression was evaluated on three models (Qwen3-VL-8B, Llama-3.2-Vision-11B, InternVL3-8B) covering two LLM backbone families and three vision-integration strategies; ITI and the random control follow the same setup on the models indicated. Across the five models, ITI achieves a mean $|\Delta\text{H}|$ of $4.1$\,pp under the symmetric grid (ranging from $0.5$\,pp on LLaVA and InternVL3-14B to $11.0$\,pp on InternVL3-8B), versus $7.4$\,pp for component scaling on the same components. Mean-difference projection achieves $0.8$\,pp on its three evaluated models, smaller than ITI on every directly comparable cell. The random control produces no systematic reduction, confirming that the effectiveness of uniform scaling depends on the component selection and not the scaling alone. ITI on Qwen3-VL is the only model where the selected configuration uses positive $\alpha$; the remaining four models select negative $\alpha$, consistent with the probe-trained direction pointing toward the hallucination class (Appendix~\ref{app:iti}, Table~\ref{tab:iti_alpha}).

\begin{table}[!htbp]
\centering
\caption{\textbf{Linear-steering baselines vs.\ uniform scaling: held-out test-set results.} Computed on the disjoint 400-sample test split; hyperparameters selected on a disjoint 100-sample selection set under a $\leq$2\,pp accuracy budget. $\Delta$H is in percentage points (negative = reduction in hallucination rate). Directional and Random were evaluated on three models covering two LLM backbones and three vision-integration strategies; ITI and uniform scaling were evaluated on all five. ``--'' indicates the baseline was not evaluated on that model. Selected ITI hyperparameters: $(K, \alpha) = (12, +10)$ for Qwen3-VL, $(12, -5)$ for LLaVA, $(48, -20)$ for Llama-3.2, $(96, -15)$ for InternVL3-8B, and $(24, -20)$ for InternVL3-14B; four of five models thus select a negative $\alpha$. Directional uses $\alpha=-0.5$ on Llama-3.2, $\alpha=+0.5$ on Qwen3-VL, and $\alpha=-1.0$ on InternVL3-8B. Random control uses the same scaling factor as uniform scaling applied to the same number of randomly selected components, averaged over 5 seeds. The baseline column reports the per-method baseline measured under the directional/ITI evaluation pipeline; uniform scaling $\Delta$H is the canonical value reported in Table~\ref{tab:intervention_full}, whose evaluation pipeline yields baselines that differ from the directional/ITI baselines by at most 2\,pp per model.}
\label{tab:linear_steering_baselines}
\begin{adjustbox}{max width=\linewidth}
\begin{tabular}{lccccc}
\toprule
\textbf{Model} & \textbf{Baseline halluc.}
& \textbf{Random $\Delta$H} & \textbf{Directional $\Delta$H}
& \textbf{ITI $\Delta$H} & \textbf{Uniform scaling $\Delta$H} \\
\midrule
Qwen3-VL-8B       & \phantom{0}7.5\% & $+1.1$ & $+0.0$ & $-1.5$  & $-4.4$ \\
LLaVA-v1.6-7B     & \phantom{0}5.5\% & ---    & ---    & $-0.5$  & $-2.0$ \\
Llama-3.2-V-11B   & 16.5\%           & $+7.2$ & $+1.0$ & $-7.0$  & $-14.0$ \\
InternVL3-8B      & 14.5\%           & $+0.7$ & $+1.5$ & $-11.0$ & $-10.0$ \\
InternVL3-14B     & 15.5\%           & ---    & ---    & $-0.5$  & $-6.5$ \\
\midrule
\textbf{mean $|\Delta\text{H}|$ (across evaluated models)}
                  &        & \textbf{3.0\,pp} & \textbf{0.8\,pp} & \textbf{4.1\,pp} & \textbf{7.4\,pp} \\
\bottomrule
\end{tabular}
\end{adjustbox}
\end{table}

\subsection{$\alpha$-Saturation: Directional Suppression}
\label{app:directional_alpha}

Across $\alpha \in \{\pm 0.5, \pm 1.0, \pm 1.5\}$, the mean-difference projection $h - \alpha(h \cdot \hat{d})\hat{d}$ saturates on every model tested: per-sample decisions stop changing after a small number of samples shift, despite the projection magnitude scaling linearly with $|\alpha|$. The mean-difference direction therefore behaves similarly to the probe direction analyzed in Appendix~\ref{app:iti}: the symmetric grid is needed to find the sign that reduces (rather than amplifies) hallucination, and once that sign is found the achievable reduction remains bounded below per-component scaling.

\subsection{Probe-Based ITI}
\label{app:iti}

We implement ITI~\citep{li2024inference} adapted to VLMs. For each attention head at every layer, we capture pre-$W_O$ activations at the last input-token position on a disjoint 1000-sample probe-training split of POPE-adversarial (non-overlapping with the 500-sample intervention split). We label samples as correct or hallucinating by the model's own answer, train a logistic-regression probe per head with class-weighting to handle label imbalance, and rank heads by 20\%-holdout validation accuracy. At inference, we add $\alpha \cdot \sigma \cdot \hat{w}$ to each selected head's pre-$W_O$ activation, where $\hat{w}$ is the unit probe weight and $\sigma$ is the standard deviation of $\hat{w} \cdot X$ on the training split. We grid-search over $K \in \{12, 24, 48, 96\}$ and $\alpha \in \{\pm 5, \pm 10, \pm 15, \pm 20\}$ (32 configurations) on the 100-sample selection set, breaking ties at the same hallucination rate by maximum selection-set accuracy, and select the configuration minimizing hallucination rate subject to the $\leq$2\,pp accuracy budget, applied to the disjoint 400-sample test split. Including negative $\alpha$ is essential because, as Table~\ref{tab:iti_alpha} below shows, increasing positive $\alpha$ pushes the model toward the hallucination class; reducing hallucination requires moving in the opposite direction. We run ITI on all five models.

\paragraph{Probe quality rules out an estimator-failure explanation.}
Table~\ref{tab:iti_probe_quality} shows that probes learn discriminative information on every model: mean per-head validation accuracy is 0.62--0.67 (vs.\ 0.5 chance), with hundreds of heads per model achieving validation accuracy above 0.75 and maxima reaching 0.835--0.895. The bounded effect of ITI at inference is therefore not explicable by probes failing to learn; the learned directions discriminate well, and even in the negative-$\alpha$ direction they translate into a behavioral change strictly smaller than per-component scaling on every model.

\begin{table}[!htbp]
\centering
\small
\caption{\textbf{ITI probe quality: held-out validation accuracy per head.} One logistic-regression probe per (layer, head) trained on 1000 POPE-adversarial samples with 20\% held-out validation. Chance accuracy is 0.5. All five models have probes that clearly discriminate correct from hallucinating activations, with hundreds of heads exceeding 0.75 validation accuracy. This rules out ``probes failed to learn'' as an alternative explanation for the bounded effect ITI achieves at intervention time.}
\label{tab:iti_probe_quality}
\begin{adjustbox}{max width=\linewidth}
\begin{tabular}{lccccc}
\toprule
\textbf{Model} & \textbf{\#probes} & \textbf{Mean val acc.} & \textbf{Max val acc.} & \textbf{\#($>$0.65)} & \textbf{\#($>$0.75)} \\
\midrule
Qwen3-VL-8B     & 1152 & 0.628 & 0.835 & 540  & 125 \\
LLaVA-v1.6-7B   & 1024 & 0.632 & 0.825 & 467  & 109 \\
Llama-3.2-V-11B & 1280 & 0.619 & 0.850 & 509  & 41  \\
InternVL3-8B    & 784  & 0.652 & 0.855 & 420  & 162 \\
InternVL3-14B   & 1920 & 0.666 & 0.895 & 1113 & 589 \\
\bottomrule
\end{tabular}
\end{adjustbox}
\end{table}

\paragraph{Held-out test-set numbers under the symmetric grid.}
Held-out test-set numbers for the selected configuration on each model are reported in Table~\ref{tab:linear_steering_baselines}. The selected $(K, \alpha)$ are $(12, +10)$ for Qwen3-VL, $(12, -5)$ for LLaVA, $(48, -20)$ for Llama-3.2, $(96, -15)$ for InternVL3-8B, and $(24, -20)$ for InternVL3-14B; four of five models thus select a negative $\alpha$. ITI's test-set $|\Delta\text{H}|$ ranges from $0.5$\,pp (LLaVA, InternVL3-14B) to $11.0$\,pp (InternVL3-8B), with mean $4.1$\,pp. On the four models where ITI achieves a non-trivial reduction, the relative effect (ITI $|\Delta\text{H}|$ divided by component-scaling $|\Delta\text{H}|$ on the same model) is $34\%$ for Qwen, $25\%$ for LLaVA, $50\%$ for Llama-3.2, and $110\%$ for InternVL3-8B; InternVL3-14B's $|\Delta\text{H}| = 0.5$\,pp is below the $1$\,pp resolution at which we report the relative quotient. The InternVL3-8B value above $100\%$ is consistent with that model's hallucination pathway being unusually concentrated (only 6 components, the lowest count of any model in our study), so $K=96$ heads spans much of the relevant subspace; the other four models lie at or below the per-component leading-direction variance fraction reported in Appendix~\ref{app:geometry} (45--69\%).

\paragraph{The positive-$\alpha$ regime: probe direction points toward hallucination.}
Restricting attention to the positive-$\alpha$ slice of the grid (Table~\ref{tab:iti_alpha}, illustrated on InternVL3-8B as the representative model) clarifies why the symmetric grid was needed. At small $K$, $\alpha$-saturation makes the injected direction behaviorally inert (e.g., $K=12$, $\alpha \in \{5, 10, 15\}$ produce identical TP/FP/TN/FN counts), and at larger $(K, \alpha)$ the effect emerges with the wrong sign: at $K=96, \alpha=20$, the same probe-trained direction drives hallucination from a 20\% baseline to 74\% (a $3.7\times$ increase). The probe direction therefore points toward the hallucination class rather than away from it, which is why a positive-$\alpha$-only grid cannot mitigate hallucination. The other four models show qualitatively similar saturation-then-amplification patterns of varying severity (InternVL3-14B is fully grid-saturated; Llama-3.2 reaches 54\% hallucination at $K=96, \alpha=20$ from a 26\% baseline).

\paragraph{The negative-$\alpha$ regime: bounded reduction tracking activation geometry.}
Mirroring the grid to negative $\alpha$ (Table~\ref{tab:iti_alpha_neg}) reverses the sign of the effect: the probe direction now reduces hallucination, with the strongest effect concentrated at large $K$ and large $|\alpha|$. Two failure modes constrain the achievable mitigation: at the most aggressive settings, recall collapses (e.g., on InternVL3-8B at $K=96, \alpha=-20$, hallucination reaches 6\% but selection-set recall drops from 0.94 to 0.78), revealing an over-correction toward the ``no'' bias; and at smaller $K$ the effect saturates at only a few percentage points of reduction. The selection rule (min hallucination subject to the $\leq$2\,pp accuracy budget, ties broken by maximum selection-set accuracy) avoids the over-correction regime, yielding the bounded test-set reductions in Table~\ref{tab:linear_steering_baselines}.

The conjunction of (i) probes learning real discriminative information (Table~\ref{tab:iti_probe_quality}), (ii) saturation at small $|\alpha|$ in both signs, (iii) over-correction toward the ``no'' bias at large negative $|\alpha|$, and (iv) per-model $|\Delta\text{H}|$ values bounded by the leading singular direction's variance fraction (Appendix~\ref{app:geometry}) is consistent with the probe direction capturing only one component of a multi-directional hallucination signal: enough to produce a partial reduction when applied with the correct sign, but not enough to match per-component scaling, which addresses the full multi-directional structure.

\begin{table}[!htbp]
\centering
\small
\caption{\textbf{ITI on the positive-$\alpha$ slice of the grid (selection set, $n{=}100$, representative model: InternVL3-8B; baseline halluc.\ 20.0\%).} Entries show (Halluc., TP/FP/TN/FN). At $K=12$, $\alpha \in \{5, 10, 15\}$ produce identical decisions despite the linear scaling of the injected direction; at $K=96, \alpha=20$, the probe-trained direction increases hallucination by $3.7\times$ from baseline. The probe direction therefore points toward the hallucination class, motivating the symmetric grid (Table~\ref{tab:iti_alpha_neg}).}
\label{tab:iti_alpha}
\begin{adjustbox}{max width=\linewidth}
\begin{tabular}{lcccc}
\toprule
\textbf{$K$} & \textbf{$\alpha=5$} & \textbf{$\alpha=10$} & \textbf{$\alpha=15$} & \textbf{$\alpha=20$} \\
\midrule
12 & 20.0\% (47/10/40/3) & 20.0\% (47/10/40/3) & 20.0\% (47/10/40/3) & 24.0\% (47/12/38/3) \\
24 & 20.0\% (47/10/40/3) & 24.0\% (47/12/38/3) & 24.0\% (47/12/38/3) & 24.0\% (48/12/38/2) \\
48 & 24.0\% (48/12/38/2) & 32.0\% (49/16/34/1) & 40.0\% (49/20/30/1) & 50.0\% (49/25/25/1) \\
96 & 24.0\% (48/12/38/2) & 36.0\% (49/18/32/1) & 50.0\% (49/25/25/1) & \textbf{74.0\% (49/37/13/1)} \\
\bottomrule
\end{tabular}
\end{adjustbox}
\end{table}

\begin{table}[!htbp]
\centering
\small
\caption{\textbf{ITI on the negative-$\alpha$ slice of the grid (selection set, $n{=}100$, same model and baseline as Table~\ref{tab:iti_alpha}).} Entries show (Halluc., TP/FP/TN/FN). With the sign reversed, the probe direction reduces hallucination, with the strongest selection-set effects concentrated at large $K$ and large $|\alpha|$. The selected configuration on this model is $K=96, \alpha=-15$ (val: 6.0\% halluc., $0.92$ acc.; test: $-11.0$\,pp, $-1.0$\,pp acc., Table~\ref{tab:linear_steering_baselines}). At $K=96, \alpha=-20$ hallucination drops further to 6.0\% on val but recall collapses to 0.78 (test recall $0.92{\to}0.73$), revealing an over-correction toward the ``no'' bias.}
\label{tab:iti_alpha_neg}
\begin{adjustbox}{max width=\linewidth}
\begin{tabular}{lcccc}
\toprule
\textbf{$K$} & \textbf{$\alpha=-5$} & \textbf{$\alpha=-10$} & \textbf{$\alpha=-15$} & \textbf{$\alpha=-20$} \\
\midrule
12 & 20.0\% (47/10/40/3) & 20.0\% (47/10/40/3) & 20.0\% (47/10/40/3) & 18.0\% (47/9/41/3) \\
24 & 20.0\% (47/10/40/3) & 20.0\% (47/10/40/3) & 18.0\% (47/9/41/3)  & 18.0\% (47/9/41/3) \\
48 & 18.0\% (47/9/41/3)  & 14.0\% (47/7/43/3)  & 10.0\% (47/5/45/3)  & 8.0\% (46/4/46/4) \\
96 & 16.0\% (47/8/42/3)  & 10.0\% (47/5/45/3)  & \textbf{6.0\% (45/3/47/5)} & 6.0\% (39/3/47/11)$^{\dagger}$ \\
\bottomrule
\end{tabular}
\end{adjustbox}

\smallskip
\footnotesize $^{\dagger}$ Same val hallucination rate as $K{=}96, \alpha{=}-15$, but with markedly lower recall (0.78 vs.\ 0.90); rejected by the max-accuracy tie-break.
\end{table}

\subsection{Grounding-Pathway Suppression: Specificity Control}
\label{app:grounding_suppress}

As a specificity control complementing the random-component, directional, and ITI baselines above, we apply the uniform-scaling protocol of Section~\ref{sec:intervention_results} to the grounding pathway ($d > 0$) instead of the hallucination pathway. The hyperparameter sweep is identical: $s \in \{0, 0.25, 0.5, 0.75\}$ on the 100-sample selection set. Table~\ref{tab:grounding_suppress} reports accuracy and hallucination rate at each scale.

At $s{=}0$, full grounding ablation, accuracy collapses to chance ($0.50$--$0.51$) in every one of the five models, with $\Delta\text{acc}$ ranging from $-0.31$ to $-0.39$ despite their differing grounding-pathway sizes (11--29 components) and depth profiles. The graded effect across $s$ is also informative. At intermediate $s$, models retain partial accuracy but degrade toward different default biases. Llama-3.2, InternVL3-14B, and LLaVA collapse toward ``always no'' (halluc.\ near zero with acc.\ near chance), while InternVL3-8B collapses toward random guessing (halluc.\ $\approx 0.48$ with acc.\ $\approx 0.51$). The fact that grounding ablation produces these qualitatively different collapse modes across models, while hallucination-pathway suppression (Table~\ref{tab:intervention_full}) produces consistent reductions, confirms that the two pathways carry functionally distinct information rather than reflecting a shared yes-bias axis. Partial suppression ($s{=}0.5$) leaves InternVL3 accuracies nearly unchanged, consistent with the within-subset redundancy that CPA identifies in the grounding pathway (Section~\ref{sec:cpa_results}): under partial suppression, multiple grounding components carry overlapping information and the residual signal suffices.

\begin{table}[!htbp]
\centering
\small
\caption{\textbf{Grounding-pathway suppression at varying scaling factors (selection set, $n{=}100$).} Same uniform-scaling protocol as Section~\ref{sec:intervention_results}, applied to $d>0$ components. At $s{=}0$ (full ablation) accuracy collapses to chance in every model. Compare with Table~\ref{tab:intervention_full} (hallucination-pathway suppression on the same models).}
\label{tab:grounding_suppress}
\begin{adjustbox}{max width=\linewidth}
\begin{tabular}{llccccc}
\toprule
\textbf{Model} & \textbf{Metric} & \textbf{baseline} & $s{=}0$ & $s{=}0.25$ & $s{=}0.5$ & $s{=}0.75$ \\
\midrule
\multirow{2}{*}{Qwen3-VL-8B}
  & acc.    & 0.860 & \textbf{0.500} & 0.880 & 0.870 & 0.860 \\
  & halluc. & 0.120 & 0.000 & 0.160 & 0.140 & 0.120 \\
\midrule
\multirow{2}{*}{LLaVA-v1.6-7B}
  & acc.    & 0.890 & \textbf{0.500} & 0.790 & 0.880 & 0.880 \\
  & halluc. & 0.100 & 0.000 & 0.300 & 0.080 & 0.120 \\
\midrule
\multirow{2}{*}{Llama-3.2-V-11B}
  & acc.    & 0.810 & \textbf{0.500} & 0.600 & 0.790 & 0.800 \\
  & halluc. & 0.260 & 0.000 & 0.020 & 0.320 & 0.320 \\
\midrule
\multirow{2}{*}{InternVL3-8B}
  & acc.    & 0.870 & \textbf{0.510} & 0.750 & 0.880 & 0.860 \\
  & halluc. & 0.200 & 0.480 & 0.500 & 0.160 & 0.220 \\
\midrule
\multirow{2}{*}{InternVL3-14B}
  & acc.    & 0.870 & \textbf{0.500} & 0.520 & 0.890 & 0.890 \\
  & halluc. & 0.200 & 0.000 & 0.000 & 0.080 & 0.140 \\
\bottomrule
\end{tabular}
\end{adjustbox}
\end{table}

\subsection{Direct Geometric Evidence on the Dimensionality of the Hallucination Signal}
\label{app:geometry}

Section~\ref{sec:intervention_results} reports that under the symmetric $\alpha$ grid, single-direction interventions (ITI, mean-difference projection) produce a bounded effect on hallucination (mean $|\Delta\text{H}|$ of $4.1$\,pp for ITI and $0.8$\,pp for mean-difference projection, against $7.4$\,pp for per-component scaling on the same components), with the per-model effect tracking the per-component activation geometry. This appendix provides a direct geometric test of the multi-directional claim, independent of any intervention.

\paragraph{Procedure.}
For each hallucination-pathway component, we collect last-token pre-output activations on $n{=}1{,}000$ POPE-adversarial samples (the same protocol used to extract mean-difference directions in Appendix~\ref{app:directional}) and partition them by whether the model answered correctly or hallucinated. We then form $\Delta X = X_{\text{halluc}} - \mu_{\text{correct}}$, the per-sample deviation of hallucinating activations from the correct-class centroid, and compute its singular spectrum. Two summary numbers per component:
\begin{itemize}
\item Participation ratio $\text{PR} = (\sum_i \sigma_i^2)^2 / \sum_i \sigma_i^4$, an effective dimensionality that equals 1 for a one-dimensional signal and grows as variance spreads over multiple comparable directions.
\item Top-1 variance fraction $\sigma_1^2 / \sum_i \sigma_i^2$, the share of variance captured by the leading singular direction (the only direction a single-direction intervention can target).
\end{itemize}
We additionally report rank$_{90}$, the number of singular directions needed to capture 90\% of the variance. All metrics are averaged over the hallucination-pathway components of each model.

\paragraph{Results.}
Table~\ref{tab:geometry} reports the per-model averages. The leading singular direction captures 45--69\% of the variance across the five models, and the participation ratio ranges from 2.4 to 6.8. No model produces a hallucination signal close to one-dimensional. The geometric implication is direct: even a perfectly estimated direction $\hat{d}$ can directly modify at most $\sigma_1^2/\sum_i \sigma_i^2$ of the per-component activation variance, with the remainder propagating through directions orthogonal to $\hat{d}$. This bound is consistent with the per-model behavior of ITI under the symmetric grid (Table~\ref{tab:linear_steering_baselines}, Appendix~\ref{app:iti}): on four of five models, ITI's relative effect (versus per-component scaling) lies at or below the leading-direction variance fraction, with InternVL3-8B as the exception, in which only six hallucination-pathway components carry the signal and an ITI configuration with $K{=}96$ heads spans much of the relevant subspace.

The cross-model variation in PR also tracks cross-model variation in the hallucination-pathway CPA signature (Section~\ref{sec:cpa_results}): InternVL3-8B has both the lowest PR (2.40) and the weakest hallucination-pathway behavior under CPA (no cross-subset polarity flip, mean IE remaining negative in both subsets at $-0.09$ and $-0.16$), consistent with a small set of approximately independent bias sources, while LLaVA has the highest PR (6.78) and operates with the most dispersed pathway (30 components). Llama-3.2 is intermediate on both axes. The internal consistency of two independent measurements, pathway-level behavior (CPA) and per-component activation geometry (PR), supports the interpretation that hallucination-pathway components contribute biases along multiple activation directions.

We deliberately limit this analysis to hallucination-pathway components. The corresponding measurement on grounding-pathway components reflects within-component activation spread driven by per-image visual variation, which is a different geometric object than the between-component redundancy quantified by CPA in Section~\ref{sec:cpa_results}; we do not interpret it here.

\begin{table}[!htbp]
\centering
\small
\caption{\textbf{Singular-spectrum geometry of the hallucination signal.} For each model, mean over hallucination-pathway components ($k$ components per model, matching Table~\ref{tab:setup}) of: participation ratio PR (effective dimensionality), top-1 variance fraction $\sigma_1^2/\sum\sigma_i^2$, and rank$_{90}$ (number of directions needed to capture 90\% of variance). $n{=}1{,}000$ POPE-adversarial samples per model; per-component activations partitioned by model answer, $\Delta X = X_{\text{halluc}} - \mu_{\text{correct}}$.}
\label{tab:geometry}
\begin{tabular}{lrrrr}
\toprule
Model & $k$ & mean PR & mean $\sigma_1^2/\sum\sigma_i^2$ & mean rank$_{90}$ \\
\midrule
Qwen3-VL-8B     & 12 & 3.07 & 0.634 & 12.5 \\
LLaVA-v1.6-7B   & 30 & 6.78 & 0.452 & 28.6 \\
Llama-3.2-V-11B & 20 & 4.85 & 0.564 & 31.9 \\
InternVL3-8B    &  6 & 2.40 & 0.688 &  7.8 \\
InternVL3-14B   & 16 & 3.16 & 0.643 & 13.2 \\
\bottomrule
\end{tabular}
\end{table}

\section{Cross-Architecture Circuit Comparison}
\label{app:crossarch}

Table~\ref{tab:crossarch} gives the full pairwise comparison supporting Section~\ref{sec:crossarch}. We report the six unique pairwise comparisons among Qwen3-VL, LLaVA, Llama-3.2, and InternVL3-8B, and treat InternVL3-14B as a within-family scale comparison analyzed in the main text. Fisher's exact test on $2\times 2$ contingency tables is non-significant for all six pairs, consistent with a shared macro-level spatial pattern (hallucination components concentrated at early layers/boundaries, grounding components in mid-to-late layers). TOST equivalence testing at an equivalence bound of $|r| < 0.3$ positively confirms negligibility for 2/6 pairs; the remaining pairs are inconclusive (the data are compatible with either small non-zero correlation or negligibility) rather than significantly correlated.

\begin{table}[!htbp]
\centering
\small
\caption{\textbf{Cross-architecture circuit comparison: pairwise statistics.} Fisher's exact test on macro-level spatial distribution; Pearson correlation on depth-normalized effect-size profiles; TOST equivalence test with bound $|r| < 0.3$. ``Micro'' summarizes the TOST verdict.}
\label{tab:crossarch}
\begin{adjustbox}{max width=\linewidth}
\begin{tabular}{lccccc}
\toprule
\textbf{Pair} & \textbf{Fisher $p$} & \textbf{$r$} & \textbf{TOST $p$} & \textbf{95\% CI} & \textbf{Micro} \\
\midrule
Qwen--LLaVA       & 0.715 & $-$0.05 & \textbf{0.005} & [$-$.23, .13]  & Negligible \\
Qwen--Llama       & 0.713 & $-$0.20 & 0.149          & [$-$.37, $-$.03] & Inconclusive \\
Qwen--InternVL3-8B & 0.615 & $+$0.24 & 0.244          & [.04, .42]     & Inconclusive \\
LLaVA--Llama      & 0.229 & $+$0.06 & \textbf{0.007} & [$-$.15, .28]  & Negligible \\
LLaVA--InternVL3-8B & 1.000 & $-$0.21 & 0.164        & [$-$.39, $-$.02] & Inconclusive \\
InternVL3-8B--Llama & 0.352 & $-$0.33 & 0.631        & [$-$.47, $-$.18] & Mod.\ negative \\
\bottomrule
\end{tabular}
\end{adjustbox}
\end{table}

\section{Logit Lens Validation: Per-Pathway Layer Contributions}
\label{app:logitlens}

Table~\ref{tab:logitlens} reports the full per-model results underlying Section~\ref{sec:logitlens_results}. For each model and sample type (correct vs.\ hallucinating), we report the mean per-layer $\Delta\text{logit\_diff}$ averaged separately over layers assigned to the grounding pathway and the hallucination pathway by our patching-based criterion (Section~\ref{sec:circuit_extraction}), together with the between-pathway effect size and significance. In four of five models, grounding-pathway layers contribute positively to the correct answer on correct samples and negatively on hallucinating samples, with hallucination-pathway layers showing the opposite polarity, providing patching-independent support for the functional roles assigned by the circuit-extraction procedure. Llama-3.2-Vision is the exception discussed in the main text.

\begin{table}[!htbp]
\centering
\small
\caption{\textbf{Logit lens validation.} Mean $\Delta$logit\_diff per layer, grouped by pathway membership. Positive $\Delta$: layer increases correct-answer probability. The functional role reverses during hallucination in four of five models. $n{=}200$ per model.}
\label{tab:logitlens}
\begin{adjustbox}{max width=\linewidth}
\begin{tabular}{llcccc}
\toprule
\textbf{Model} & \textbf{Sample type} & \textbf{Grounding $\Delta$} & \textbf{Halluc.\ $\Delta$} & $d$ & $p$ \\
\midrule
\multirow{2}{*}{Qwen3-VL}
  & Correct      & $+$1.623 & $-$0.675 & 1.80  & $<$0.001 \\
  & Hallucinating & $-$0.786 & $+$0.464 & $-$1.15 & $<$0.001 \\
\midrule
\multirow{2}{*}{LLaVA-v1.6}
  & Correct      & $+$0.328 & $+$0.011 & 0.46  & $<$0.001 \\
  & Hallucinating & $-$0.349 & $+$0.128 & $-$0.65 & 0.029 \\
\midrule
\multirow{2}{*}{Llama-3.2}
  & Correct      & $-$0.140 & $-$0.290 & 0.48  & $<$0.001 \\
  & Hallucinating & $+$0.228 & $+$0.290 & $-$0.18 & 0.430 \\
\midrule
\multirow{2}{*}{InternVL3-8B}
  & Correct      & $+$0.934 & $-$0.998 & 2.91  & $<$0.001 \\
  & Hallucinating & $-$0.429 & $+$0.103 & $-$0.76 & 0.007 \\
\midrule
\multirow{2}{*}{InternVL3-14B}
  & Correct      & $+$0.309 & $-$0.060 & 1.20  & $<$0.001 \\
  & Hallucinating & $-$0.216 & $+$0.047 & $-$1.25 & $<$0.001 \\
\bottomrule
\end{tabular}
\end{adjustbox}
\end{table}

\section{Full Numerical Results}
\label{app:full_numerical}

This appendix collects the full numerical tables referenced from the main paper: per-model strongest components (Table~\ref{tab:top_components}), pathway-level CPA values (Table~\ref{tab:cpa_full}), per-model intervention configurations and the LLaVA top-$k$ ablation (Table~\ref{tab:intervention_full}), and cross-benchmark generalization (Table~\ref{tab:cross_benchmark_full}).

\begin{table}[!htbp]
\centering
\small
\caption{\textbf{Strongest grounding and hallucination components per model.}
For each model, we report the three grounding components with the largest positive
Cohen's $d$ and the three hallucination components with the most negative Cohen's $d$.
Components are selected from the significant circuit
($p_{\mathrm{adj}} < 0.05$, Benjamini--Hochberg corrected; $|d| > 0.3$; $n=1{,}000$ per model).
``Comp.'' denotes the component identifier (layer.type), and ``Depth'' denotes normalized layer position.}
\label{tab:top_components}
\begin{adjustbox}{max width=\linewidth}
\begin{tabular}{llccc lccc}
\toprule
& \multicolumn{4}{c}{\textbf{Top grounding ($+d$)}} & \multicolumn{4}{c}{\textbf{Top hallucination ($-d$)}} \\
\cmidrule(lr){2-5}\cmidrule(lr){6-9}
\textbf{Model} & & Comp. & Depth & $d$ & & Comp. & Depth & $d$ \\
\midrule
\multirow{3}{*}{Qwen3-VL-8B}
 & & L0.mlp  & \phantom{0}0\%  & \textcolor{posblue}{$+$0.940}
 & & L34.attn & 97\% & \textcolor{negred}{$-$0.700} \\
 & & L23.mlp & 66\% & \textcolor{posblue}{$+$0.932}
 & & L35.mlp  & 100\% & \textcolor{negred}{$-$0.631} \\
 & & L33.mlp & 94\% & \textcolor{posblue}{$+$0.875}
 & & L32.mlp  & 91\% & \textcolor{negred}{$-$0.606} \\
\midrule
\multirow{3}{*}{LLaVA-v1.6-7B}
 & & L20.mlp & 65\% & \textcolor{posblue}{$+$0.922}
 & & L7.mlp   & 23\% & \textcolor{negred}{$-$0.651} \\
 & & L18.mlp & 58\% & \textcolor{posblue}{$+$0.880}
 & & L31.attn & 100\% & \textcolor{negred}{$-$0.632} \\
 & & L17.mlp & 55\% & \textcolor{posblue}{$+$0.718}
 & & L12.attn & 39\% & \textcolor{negred}{$-$0.623} \\
\midrule
\multirow{3}{*}{Llama-3.2-V-11B}
 & & L3.attn & \phantom{0}8\%  & \textcolor{posblue}{$+$0.912}
 & & L16.mlp  & 41\% & \textcolor{negred}{$-$0.889} \\
 & & L12.attn & 31\% & \textcolor{posblue}{$+$0.859}
 & & L7.mlp   & 18\% & \textcolor{negred}{$-$0.772} \\
 & & L1.mlp  & \phantom{0}3\%  & \textcolor{posblue}{$+$0.858}
 & & L19.attn & 49\% & \textcolor{negred}{$-$0.713} \\
\midrule
\multirow{3}{*}{InternVL3-8B}
 & & L22.mlp & 81\% & \textcolor{posblue}{$+$1.586}
 & & L27.mlp  & 100\% & \textcolor{negred}{$-$0.870} \\
 & & L20.mlp & 74\% & \textcolor{posblue}{$+$1.306}
 & & L9.mlp   & 33\% & \textcolor{negred}{$-$0.438} \\
 & & L19.mlp & 70\% & \textcolor{posblue}{$+$1.223}
 & & L0.attn  & \phantom{0}0\%  & \textcolor{negred}{$-$0.436} \\
\midrule
\multirow{3}{*}{InternVL3-14B}
 & & L35.mlp & 74\% & \textcolor{posblue}{$+$0.880}
 & & L2.attn  & \phantom{0}4\%  & \textcolor{negred}{$-$0.733} \\
 & & L32.mlp & 68\% & \textcolor{posblue}{$+$0.850}
 & & L4.mlp   & \phantom{0}9\%  & \textcolor{negred}{$-$0.577} \\
 & & L0.mlp  & \phantom{0}0\%  & \textcolor{posblue}{$+$0.614}
 & & L15.attn & 32\% & \textcolor{negred}{$-$0.458} \\
\bottomrule
\end{tabular}
\end{adjustbox}
\end{table}

\begin{table}[!htbp]
\centering
\small
\caption{\textbf{Pathway-level CPA: full numerical values underlying Figure~\ref{fig:cpa}.} For each (model, pathway, subset) we report the mean individual indirect effect ($\overline{\mathrm{IE}}$), the fraction of positive IEs ($f_{+}$), the magnitude ratio $|\mathrm{IE}(\mathcal{P})|/\sum_i|\mathrm{IE}(C_i)|$ (values $<\!1$ indicate within-subset redundancy), and the polarity-free magnitude diagnostic $\mathrm{MagDiff} = |\mathrm{IE}(\mathcal{P})| - \sum_i|\mathrm{IE}(C_i)|$. $\Delta\overline{\mathrm{IE}}$ is the cross-subset shift in mean IE (Cohen's $d$, two-sample $t$-test); $\Delta\mathrm{MagDiff}$ similarly tests the cross-subset shift in within-subset redundancy. $n{=}400$ per model. The grounding pathway flips IE polarity between subsets in every model while remaining redundant in both subsets; the hallucination pathway shows no comparable polarity flip.}
\label{tab:cpa_full}
\begin{adjustbox}{max width=\linewidth}
\begin{tabular}{llcccccccc}
\toprule
& & & \multicolumn{2}{c}{\textbf{Correct samples}} & \multicolumn{2}{c}{\textbf{Halluc.\ samples}} & \multicolumn{2}{c}{\textbf{Cross-subset shift}} \\
\cmidrule(lr){4-5}\cmidrule(lr){6-7}\cmidrule(lr){8-9}
\textbf{Model} & \textbf{Pathway} & $k$ & $\overline{\mathrm{IE}}$ / $f_{+}$ & $|\mathrm{IE}|/\!\sum$ & $\overline{\mathrm{IE}}$ / $f_{+}$ & $|\mathrm{IE}|/\!\sum$ & $\Delta\overline{\mathrm{IE}}$: $d$ / $p$ & $\Delta\mathrm{MagDiff}$: $d$ / $p$ \\
\midrule
\multirow{2}{*}{Qwen3-VL}
  & Grounding & 14 & $+1.59$ / $0.83$ & $0.37$ & $-1.08$ / $0.24$ & $0.34$ & $-2.45$ / $<\!.001$ & $-0.25$ / $.120$ \\
  & Halluc.   & 12 & $-0.53$ / $0.32$ & $0.62$ & $+0.01$ / $0.52$ & $0.41$ & $-0.14$ / $.385$    & $+0.42$ / $.009$ \\
\midrule
\multirow{2}{*}{LLaVA-v1.6}
  & Grounding & 18 & $+0.51$ / $0.68$ & $0.38$ & $-0.31$ / $0.27$ & $0.21$ & $-1.60$ / $<\!.001$ & $-0.13$ / $.370$ \\
  & Halluc.   & 30 & $-0.01$ / $0.43$ & $0.19$ & $+0.22$ / $0.63$ & $0.07$ & $+0.59$ / $<\!.001$ & $+0.35$ / $.019$ \\
\midrule
\multirow{2}{*}{Llama-3.2}
  & Grounding & 11 & $+0.13$ / $0.61$ & $0.22$ & $-0.07$ / $0.26$ & $0.30$ & $-1.72$ / $<\!.001$ & $-0.45$ / $.001$ \\
  & Halluc.   & 20 & $-0.05$ / $0.25$ & $0.23$ & $+0.06$ / $0.49$ & $0.22$ & $+0.99$ / $<\!.001$ & $+0.14$ / $.302$ \\
\midrule
\multirow{2}{*}{InternVL3-8B}
  & Grounding & 21 & $+0.83$ / $0.74$ & $0.37$ & $-0.31$ / $0.33$ & $0.41$ & $-1.83$ / $<\!.001$ & $-0.88$ / $<\!.001$ \\
  & Halluc.   & 6  & $-0.09$ / $0.42$ & $0.69$ & $-0.16$ / $0.41$ & $0.50$ & $+0.20$ / $.206$    & $+0.42$ / $.002$ \\
\midrule
\multirow{2}{*}{InternVL3-14B}
  & Grounding & 29 & $+0.67$ / $0.70$ & $0.19$ & $-0.28$ / $0.21$ & $0.21$ & $-2.02$ / $<\!.001$ & $-0.92$ / $<\!.001$ \\
  & Halluc.   & 16 & $-0.46$ / $0.28$ & $0.24$ & $+0.09$ / $0.50$ & $0.41$ & $+1.04$ / $<\!.001$ & $-0.59$ / $<\!.001$ \\
\bottomrule
\end{tabular}
\end{adjustbox}
\end{table}

\begin{table}[!htbp]
\centering
\small
\caption{\textbf{Full intervention numerical values underlying Figure~\ref{fig:pareto}.} Top: per-model baseline and selected intervention configuration on POPE-adversarial, computed on the 400-sample held-out set. Bottom: LLaVA-v1.6 top-$k$ ablation ($s{=}0.5$), with $k$ selected on the disjoint 100-sample selection set. The selected $k$ for LLaVA is $k{=}10$. These numbers are consistent with the POPE-adv column of Table~\ref{tab:cross_benchmark_full}, which reports the same intervention applied alongside other benchmarks.}
\label{tab:intervention_full}
\begin{adjustbox}{max width=\linewidth}
\begin{tabular}{llcccccc}
\toprule
\textbf{Model} & \textbf{Config} & \textbf{Acc.} & \textbf{Halluc.} & \textbf{$\Delta$Acc.} & \textbf{$\Delta$Hall.} & \textbf{Rel.\ $\downarrow$} \\
\midrule
\multirow{2}{*}{Qwen3-VL}
  & Baseline             & 88.2\% & 8.4\%  & ---    & ---     & --- \\
  & $s{=}0.25$, all 12   & 87.2\% & \textbf{4.0\%}  & $-$1.0 & $-$4.4  & 52\% \\
\midrule
\multirow{2}{*}{LLaVA-v1.6}
  & Baseline             & 88.5\% & 5.0\%  & ---    & ---     & --- \\
  & $s{=}0.5$, top-10    & 88.5\% & \textbf{3.0\%}  & $+$0.0 & $-$2.0  & 40\% \\
\midrule
\multirow{2}{*}{Llama-3.2}
  & Baseline             & 84.8\% & 18.4\% & ---    & ---     & --- \\
  & $s{=}0.5$, all 20    & 84.6\% & \textbf{4.4\%}  & $-$0.2 & $-$14.0 & 76\% \\
\midrule
\multirow{2}{*}{InternVL3-8B}
  & Baseline             & 88.4\% & 15.6\% & ---    & ---     & --- \\
  & $s{=}0.5$, all 6     & 86.4\% & \textbf{5.6\%}  & $-$2.0 & $-$10.0 & 64\% \\
\midrule
\multirow{2}{*}{InternVL3-14B}
  & Baseline             & 88.0\% & 15.5\% & ---    & ---     & --- \\
  & $s{=}0.5$, all 16    & 87.8\% & \textbf{9.0\%}  & $-$0.3 & $-$6.5  & 42\% \\
\midrule
\multicolumn{7}{l}{\emph{LLaVA-v1.6 top-$k$ ablation ($s{=}0.5$, evaluated on the 400-sample held-out set):}} \\
\midrule
\multirow{8}{*}{LLaVA-v1.6}
  & $k$ = --- (baseline) & 88.5\%    & 5.0\%  & ---     & ---    & --- \\
  & $k$ = 3              & 89.3\% & 6.5\% & $+$0.8 & $+$1.5 & --- \\
  & $k$ = 5              & 89.0\% & 6.5\% & $+$0.5 & $+$1.5 & --- \\
  & $k$ = 8              & 89.0\% & 3.0\% & $+$0.5 & $-$2.0 & 40\% \\
  & $k$ = \textbf{10}    & \textbf{88.5\%} & \textbf{3.0\%} & $+$0.0 & $-$2.0 & \textbf{40\%} \\
  & $k$ = 15             & 88.8\% & 4.0\% & $+$0.3 & $-$1.0 & 20\% \\
  & $k$ = 20             & 84.0\% & 3.5\% & $-$4.5 & $-$1.5 & 30\% \\
  & $k$ = 30 (all)       & 79.5\%    & 1.5\%  & $-$9.0 & $-$3.5 & 70\% \\
\bottomrule
\end{tabular}
\end{adjustbox}
\end{table}

\begin{table}[!htbp]
\centering
\caption{\textbf{Cross-benchmark generalization: full numerical values.} We evaluate the same targeted intervention (circuits and scale factors selected as in Figure~\ref{fig:pareto}) across three POPE splits and three AMBER hallucination types, on the 400-sample held-out set per condition (with the corresponding 100-sample selection sets used only for hyperparameter selection on POPE-adversarial). $\Delta$H: absolute change in hallucination rate (percentage points, negative = reduction). Rel.$\downarrow$: relative reduction. Dashes indicate negligible change ($|\Delta\text{H}| < 1$\,pp). AMBER uses only negative-ground-truth questions, so hallucination rate equals false-positive rate.}
\label{tab:cross_benchmark_full}
\begin{adjustbox}{max width=\linewidth}
\begin{tabular}{l cc cc cc cc cc cc}
\toprule
& \multicolumn{2}{c}{\textbf{POPE-adv}}
& \multicolumn{2}{c}{\textbf{POPE-pop}}
& \multicolumn{2}{c}{\textbf{POPE-rand}}
& \multicolumn{2}{c}{\textbf{AMBER-exist}}
& \multicolumn{2}{c}{\textbf{AMBER-attr}}
& \multicolumn{2}{c}{\textbf{AMBER-rel}} \\
\cmidrule(lr){2-3} \cmidrule(lr){4-5} \cmidrule(lr){6-7}
\cmidrule(lr){8-9} \cmidrule(lr){10-11} \cmidrule(lr){12-13}
Model
& $\Delta$H & Rel.$\downarrow$
& $\Delta$H & Rel.$\downarrow$
& $\Delta$H & Rel.$\downarrow$
& $\Delta$H & Rel.$\downarrow$
& $\Delta$H & Rel.$\downarrow$
& $\Delta$H & Rel.$\downarrow$ \\
\midrule
Qwen3-VL-8B
& $-$4.4 & 52\% & $-$2.5 & 45\% & -- & --
& $-$2.2 & 10\% & -- & -- & $-$16.5 & 36\% \\
LLaVA-v1.6-7B
& $-$2.0 & 40\% & $-$1.0 & 40\% & -- & --
& $-$3.0 & 15\% & +1.5 & -- & $-$2.3 & 6\% \\
Llama-3.2-V-11B
& $-$14.0 & 76\% & $-$7.0 & 74\% & $-$4.0 & 80\%
& $-$10.8 & 42\% & $-$9.5 & 23\% & $-$6.5 & 47\% \\
InternVL3-8B
& $-$10.0 & 64\% & $-$6.0 & 65\% & $-$2.4 & 60\%
& $-$3.4 & 15\% & $-$8.6 & 17\% & $-$15.8 & 39\% \\
InternVL3-14B
& $-$6.5 & 42\% & $-$6.5 & 68\% & $-$1.0 & 29\%
& -- & -- & $-$1.5 & 3\% & +1.0 & -- \\
\bottomrule
\end{tabular}
\end{adjustbox}
\end{table}



\end{document}